\documentclass[conference]{IEEEtran}
\IEEEoverridecommandlockouts
\usepackage{cite}
\usepackage{amsmath,amssymb,amsfonts}
\usepackage{algorithmic}
\usepackage{graphicx}
\usepackage{textcomp}
\usepackage{xcolor}

\usepackage{epsfig}
\usepackage{graphicx}
\usepackage{amsmath}
\usepackage{amssymb}
\usepackage{booktabs,multirow}
\usepackage{colortbl}
\usepackage{xcolor}
\usepackage{pifont}
\usepackage{booktabs}
\usepackage[ruled,vlined,linesnumbered,noend]{algorithm2e}
\DontPrintSemicolon
\SetKwInput{KwIn}{Input}
\SetKwInput{KwOut}{Output}
\SetKwInput{KwParam}{Parameter}

\usepackage{todonotes}
\usepackage{multirow}
\usepackage{ulem}        
\usepackage{algorithm2e}

\usepackage{subcaption}
\usepackage{overpic}
\definecolor{headgray}{RGB}{235,235,235}
\definecolor{rowblue}{RGB}{230,244,255}

\newcommand{\cmark}{\ding{51}} 
\newcommand{\xmark}{\ding{55}} 

\usepackage{bm}

\usepackage{siunitx} 
\usepackage[pagebackref=false,breaklinks=true,letterpaper=true,colorlinks=false,bookmarks=false,citecolor=blue,linkcolor=blue]{hyperref}

\usepackage[capitalise]{cleveref}
\crefname{section}{Sect.}{Sects.}
\Crefname{section}{Sect.}{Sects.}
\crefname{figure}{Fig.}{Figs.}
\Crefname{figure}{Fig.}{Figs.}
\crefname{table}{Tab.}{Tabs.}
\Crefname{table}{Tab.}{Tabs.}

\newcommand{\myPara}[1]{\vspace{.0in}\textbf{#1}\ }

\def\BibTeX{{\rm B\kern-.05em{\sc i\kern-.025em b}\kern-.08em
    T\kern-.1667em\lower.7ex\hbox{E}\kern-.125emX}}
    
\begin{document}

\title{Camera-Aware Cross-View Alignment for Referring 3D Gaussian Splatting Segmentation}

\author{
Yuwen Tao\textsuperscript{1}, Kanglei Zhou\textsuperscript{2,$\dagger$}, Xin Tan\textsuperscript{1}, Yuan Xie\textsuperscript{1}\\
\textsuperscript{1} East China Normal University \qquad \textsuperscript{2} Tsinghua University
}

\twocolumn[{%
\renewcommand\twocolumn[1][]{#1}%
\maketitle
\centering
\includegraphics[width=0.9\linewidth,clip,trim=0 100 0 70]
{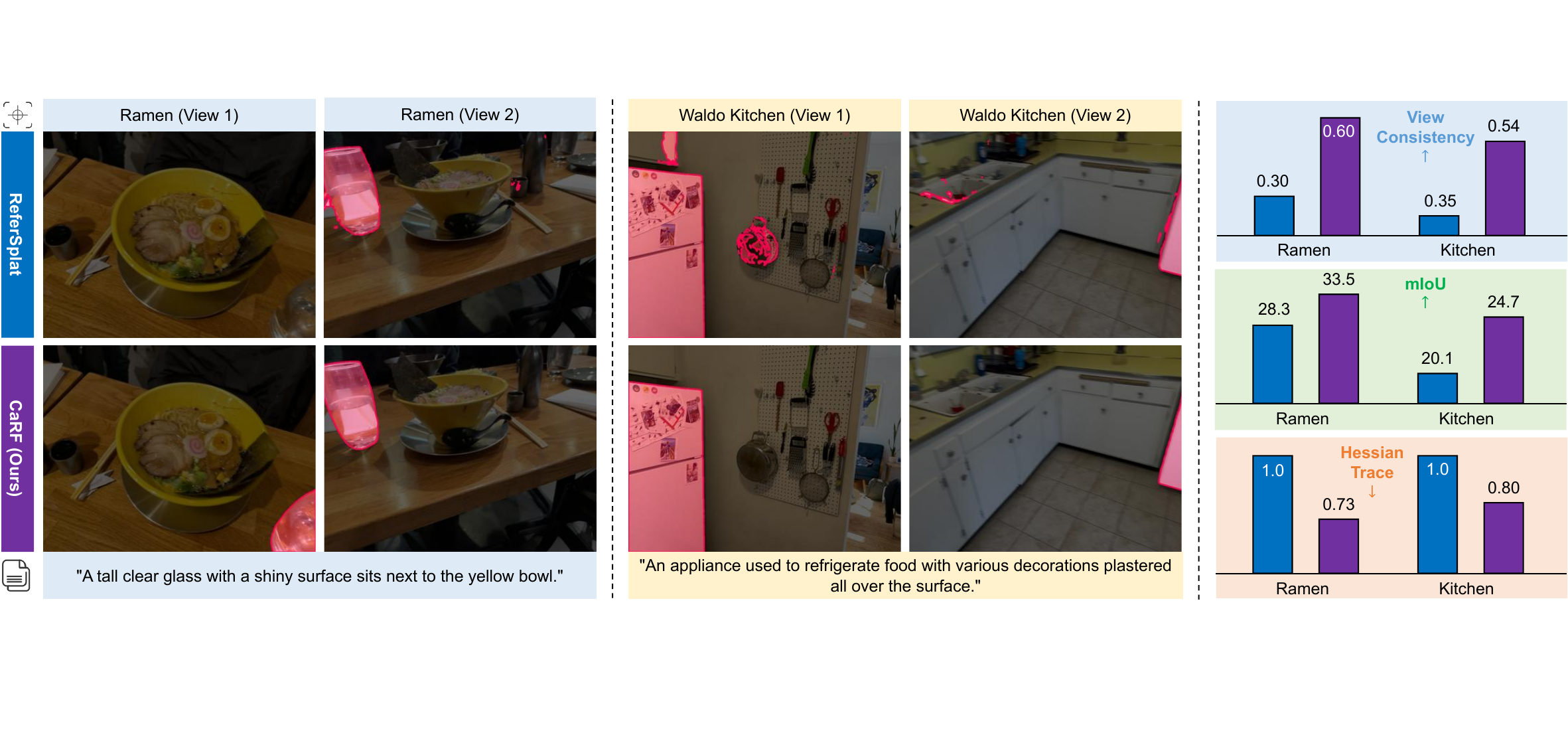}
\captionof{figure}{
Comparison between \textbf{\textcolor[HTML]{0070C0}{ReferSplat}}~\cite{ReferSplat} and our \textbf{\textcolor[HTML]{7030A0}{CaRF}} on \textit{Ramen} and \textit{Waldo Kitchen}. 
CaRF produces more geometrically consistent and accurate segmentations across views, accompanied by smaller Hessian traces, indicating improved optimization stability enabled by camera-aware cross-view alignment. 
}
\vspace{1em}
\label{fig:teaser0}
}]

\renewcommand{\thefootnote}{}
\footnotetext{$^\dagger$ Corresponding author: {\tt zhoukanglei@tsinghua.edu.cn}}

\begin{abstract}
Referring 3D Gaussian Splatting Segmentation (R3DGS) aims to ground free-form language queries in 3D Gaussian fields. 
However, existing methods rely on single-view pseudo supervision, leading to \emph{viewpoint drift} and inconsistent predictions across views. 
We propose \textbf{CaRF} (\textbf{Camera-aware Referring Field}), a camera-aware cross-view alignment framework for view-consistent referring in 3D Gaussian splatting. 
CaRF introduces \textbf{Camera-conditioned Alignment Modulation (CAM)} to inject camera geometry into Gaussian--text interactions, and \textbf{Gaussian-level Cross-view Logit Alignment (GCLA)} to explicitly align referring responses of the same Gaussians across calibrated views during training. 
By turning cross-view discrepancy into an optimizable objective, CaRF enables geometry-aware and view-consistent reasoning directly in the Gaussian space. 
Extensive experiments on three benchmarks demonstrate that CaRF achieves state-of-the-art performance, improving mIoU by \textbf{16.8\%, 4.3\%, and 2.0\%} on Ref-LERF, LERF-OVS, and 3D-OVS, respectively. 
Our code is available at \url{https://github.com/eR3R3/CaRF}.
\end{abstract}


\section{Introduction}
\label{sec:intro}

Referring 3D Gaussian Splatting Segmentation (R3DGS)~\cite{ReferSplat}, built on 3D Gaussian Splatting (3DGS)~\cite{kerbl20233dgaussiansplattingrealtime}, aims to ground free-form language queries in 3D Gaussian fields by learning a per-scene referring field over Gaussians. 
It enables open-vocabulary 3D retrieval and segmentation from calibrated images and text, which is crucial for embodied AI~\cite{shorinwa2024splatmovermultistageopenvocabularyrobotic}, autonomous driving~\cite{jatavallabhula2023conceptfusionopensetmultimodal3d, gu2023conceptgraphsopenvocabulary3dscene}, and AR/VR~\cite{jiang2024vrgsphysicaldynamicsawareinteractive, liu2024weaklysupervised3dopenvocabulary}. 
Unlike conventional methods~\cite{Qin2024LangSplat, Zhou2024Feature3DGS, choi2024clickgaussianinteractivesegmentation3d, Li2024LangSurf}, R3DGS must interpret free-form expressions with complex attributes and spatial relations, often involving occlusions or invisible targets, making \emph{view-consistent reasoning} essential and challenging.

Recent methods extend 3DGS toward open-vocabulary understanding by distilling 2D vision--language features or lifting 2D masks into 3D space, including LangSplat~\cite{Qin2024LangSplat}, Feature3DGS~\cite{Zhou2024Feature3DGS}, GaussianGrouping~\cite{Ye2024GaussianGrouping}, and OpenGaussian~\cite{Wu2024OpenGaussian}. 
ReferSplat~\cite{ReferSplat} further introduces position-aware cross-modal interaction and Gaussian--text contrastive learning for referring segmentation. 
While multi-view consistency is broadly recognized as beneficial, explicitly enforcing cross-view agreement in R3DGS is inherently challenging: pseudo masks are view-dependent and noisy, visibility varies under occlusion, and referring queries depend on fine-grained attributes and spatial relations. 
As a result, optimizing with view-specific supervision can easily induce \emph{viewpoint drift} and lead to inconsistent predictions across views (see \cref{fig:teaser0}).

To further analyze this challenge, we observe that single-view pseudo supervision in current pipelines tends to overfit view-specific artifacts, without explicitly constraining how the same Gaussians should respond across calibrated views. 
Several works attempt to improve coherence via pre-/post-processing, such as FMLGS~\cite{Tan2025FMLGS} and OmniSeg3D~\cite{ying2024omniseg3d}, but their non-differentiable and heuristic designs make them fragile for fine-grained referring scenarios~\cite{zhu2025rethinking}. 
Alternatively, differentiable clustering methods~\cite{Ye2024GaussianGrouping, silva2024contrastivegaussianclusteringweakly, Wu2024OpenGaussian, choi2024click} promote cross-view stability through feature regularization, yet still rely mainly on 2D image-level cues and underexploit native 3D geometry~\cite{ying2024omniseg3d}. 
These observations suggest that achieving robust cross-view consistency in R3DGS is \emph{non-trivial}, calling for a fully differentiable and geometry-aware formulation directly in the 3D Gaussian space.

To this end, we propose \textbf{CaRF} (Camera-aware Referring Field), a novel camera-aware \emph{cross-view alignment} paradigm for referring 3D Gaussian splatting. 
Unlike prior pipelines that optimize view-specific supervision independently, CaRF formulates referring segmentation as \emph{Gaussian-level cross-view alignment} in the 3D space. 
Specifically, CaRF introduces Camera-conditioned Alignment Modulation (CAM) to inject camera geometry into Gaussian--text interactions, disentangling view-dependent evidence from view-invariant semantics, and Gaussian-level Cross-view Logit Alignment (GCLA) to explicitly align the referring responses of the same Gaussians across calibrated views during training. 
By turning cross-view discrepancy into an optimizable objective, CaRF enables geometry-aware and view-consistent reasoning, providing a principled solution to viewpoint drift in R3DGS.

Extensive experiments on three benchmarks show that CaRF improves mIoU by \textbf{16.8\%}, \textbf{4.3\%}, and \textbf{2.0\%} over state-of-the-art methods on Ref-LERF, LERF-OVS, and 3D-OVS, respectively, while significantly enhancing cross-view consistency.

Our contributions are three-fold:
\begin{itemize}
\item We identify \emph{viewpoint drift} in R3DGS as a key challenge, showing that view-dependent pseudo supervision leads to inconsistent referring across calibrated views.
\item We propose \textbf{CaRF}, a camera-aware referring field that reformulates referring segmentation as \emph{Gaussian-level cross-view alignment} in the underlying 3D representation.
\item We develop a fully differentiable, geometry-aware alignment mechanism that enables robust and view-consistent 3D language grounding in Gaussian fields.
\end{itemize}

\section{Related Work}

\myPara{3D Neural Representations.}
Neural Radiance Fields (NeRF)~\cite{mildenhall2020nerfrepresentingscenesneural} enable high-quality novel view synthesis but suffer from slow training and rendering due to implicit representations. 
Recent explicit representations, such as voxels and point clouds, alleviate this issue. 
3D Gaussian Splatting (3DGS)~\cite{kerbl20233dgaussiansplattingrealtime} represents scenes with anisotropic Gaussians and achieves real-time rendering via fast differentiable rasterization. 
Since then, 3DGS has been extended to various downstream tasks, including Gaussian editing~\cite{chen2023gaussianeditorswiftcontrollable3d} and 3D semantic segmentation~\cite{Qin2024LangSplat, Li2024LangSurf}. 
In this work, we build upon 3DGS to study language-guided understanding at the Gaussian level.

\myPara{3D Segmentation in Gaussian Splatting.}
Large vision--language models (VLMs) such as CLIP~\cite{radford2021learningtransferablevisualmodels} have enabled open-vocabulary 2D segmentation, motivating their extension to 3D neural representations. 
Most 3DGS-based segmentation methods follow a common paradigm: extracting semantic cues from multi-view images using foundation models and distilling or lifting them into Gaussian fields. 
One branch focuses on 2D feature distillation~\cite{Qin2024LangSplat, Ji2025FastLGS}, leveraging models such as SAM, LSeg, and related approaches~\cite{kirillov2023segment, li2022languagedrivensemanticsegmentation, ravi2024sam2segmentimages}. 
Another branch lifts 2D masks to supervise 3D Gaussians, including GaussianGrouping~\cite{Ye2024GaussianGrouping}, SAGA~\cite{Cen2025SAGA}, GaussianCut~\cite{Jain2024GaussianCut}, Click-Gaussian~\cite{choi2024clickgaussianinteractivesegmentation3d}, and OpenGaussian~\cite{Wu2024OpenGaussian}. 
Although effective for category-level open-vocabulary segmentation, these methods rely heavily on single-view 2D supervision and struggle with free-form language grounding and cross-view consistency.

\myPara{Referring 3D Gaussian Splatting Segmentation.}
Referring 3D Gaussian Splatting Segmentation (R3DGS) extends referring expression segmentation in 2D~\cite{ding2021visionlanguagetransformerquerygeneration, liu2023gresgeneralizedreferringexpression, ding2023mevislargescalebenchmarkvideo} and point-based 3D referring segmentation~\cite{he2024segpointsegmentpointcloud, wang2024gpsformerglobalperceptionlocal, wang2025taylorseriesinspiredlocalstructure} to Gaussian fields. 
ReferSplat~\cite{ReferSplat} pioneers this direction by learning per-Gaussian referring features with confidence-weighted pseudo-mask supervision and Gaussian--text contrastive learning, achieving state-of-the-art performance on Ref-LERF. 
However, its supervision remains view-specific, making it prone to inconsistent predictions across views. 
Our work targets this limitation by enforcing cross-view consistency in the 3D Gaussian space.

\section{Preliminaries: Notations \& Task Definition}

\myPara{3D Gaussian Splatting (3DGS).} 
3DGS~\cite{kerbl20233dgaussiansplattingrealtime} represents a scene as a set of $N$ anisotropic Gaussians 
$\mathcal{G}=\{G_i=(\bm{\mu}_i,\bm{\Sigma}_i,\bm{c}_i,\alpha_i)\}_{i=1}^{N}$, 
where $\bm{\mu}_i$, $\bm{\Sigma}_i$, $\bm{c}_i$, and $\alpha_i$ denote the center, covariance, color, and opacity. 
Given calibrated cameras $(\mathbf{K},[\mathbf{R}|\bm{t}])$, Gaussians are projected to the image plane and rendered by alpha compositing 
$\bm{C}(\bm{p})=\sum_{i} T_i(\bm{p})\,\alpha_i'(\bm{p})\,\bm{c}_i$, with $T_i(\bm{p})=\prod_{j<i}(1-\alpha_j'(\bm{p}))$. 
The parameters are optimized by a photometric loss $\mathcal{L}_{\text{photo}}$ for differentiable reconstruction.

\myPara{Referring 3DGS (R3DGS).} 
R3DGS~\cite{ReferSplat} augments each Gaussian with a semantic feature $\bm{f}_i\in\mathbb{R}^d$, forming a language field over Gaussians. 
Given a query $q$, a pretrained language encoder produces token embeddings $\mathbf{E}\in\mathbb{R}^{L\times d}$. 
A cross-interaction module $\phi(\cdot,\cdot)$ fuses geometry and language to obtain enhanced features 
$\bm{g}_i=\phi(\bm{f}_i,\mathbf{E})$, and the referring score is computed as 
$m_i=\sum_j \bm{g}_i^\top \bm{e}_j$. 
These scores are rendered into a 2D mask $\mathbf{M}_{\text{pred}}$, supervised by pseudo ground-truth $\mathbf{M}_{\text{gt}}$ using 
$\mathcal{L}_{\text{BCE}}=\text{BCE}(\mathbf{M}_{\text{pred}},\mathbf{M}_{\text{gt}})$. 
ReferSplat further applies an object-wise contrastive loss $\mathcal{L}_{\text{con}}=\mathrm{Con}(\bm{f}_g,\bm{e}_t)$, 
where $\bm{f}_g$ aggregates top-$\tau$ Gaussians and $\bm{e}_t$ is the sentence embedding. 
The total loss is $\mathcal{L}_{\text{BCE}}+\mathcal{L}_{\text{con}}$.

\section{CaRF: Camera-Aware Referring Field}

\begin{figure}
\centering
\includegraphics[width=\linewidth,clip,trim=0 80 0 100]{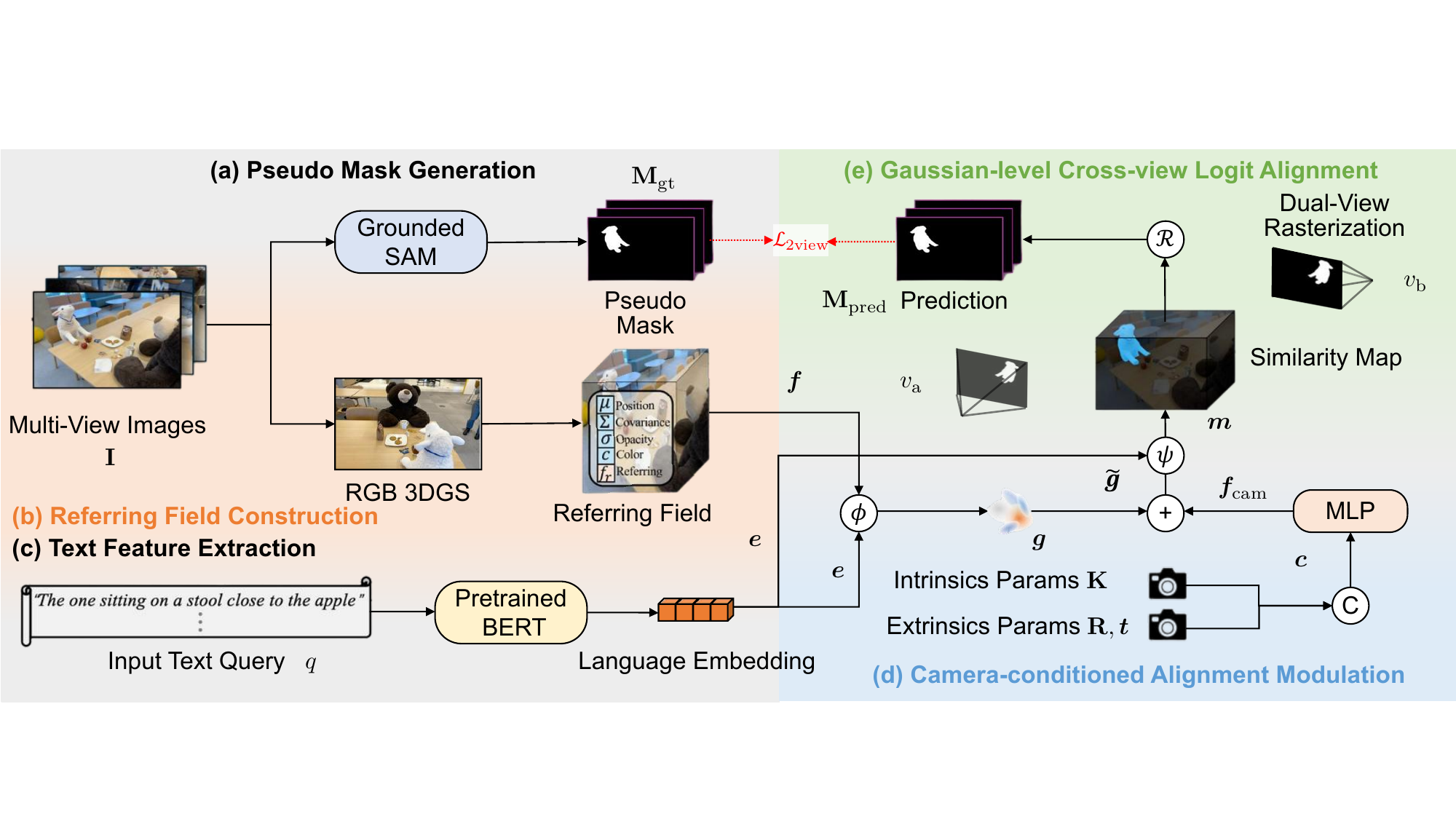}
\caption{
Overview of CaRF: CaRF integrates language interaction and camera-aware cross-view alignment in 3D Gaussian space, learning a view-consistent referring field for 3D Gaussian splatting.
}

\label{fig:framework}
\phantomsubcaption\label{fig:framework-a}
\phantomsubcaption\label{fig:framework-b}
\phantomsubcaption\label{fig:framework-c}
\phantomsubcaption\label{fig:framework-d}
\phantomsubcaption\label{fig:framework-e}
\vspace{-0.5cm}
\end{figure}

\subsection{Motivation and Framework Overview}

\myPara{Challenges.}
Despite promising progress, achieving view-consistent referring in R3DGS remains \emph{non-trivial}. 
Existing methods~\cite{ReferSplat} rely on single-view pseudo supervision, which is inherently noisy and view-dependent, making models prone to overfitting view-specific artifacts and inducing \emph{viewpoint drift}. 
As a result, the same Gaussians may yield inconsistent predictions across calibrated views (see \cref{fig:teaser0}), undermining reliable 3D spatial reasoning for free-form queries.

\myPara{Core Idea.}
We rethink referring segmentation in 3DGS as a \emph{camera-aware cross-view alignment} problem in 3D Gaussian space, where camera geometry is not used as auxiliary cues but co-defines the referring field, and cross-view consistency is enforced during feature formation.

\myPara{Framework Overview.}
\cref{fig:framework} shows the framework of CaRF. 
Given calibrated multi-view images and a free-form query, we first obtain a pseudo mask from 2D foundation models and use it as a supervisory signal. 
Each Gaussian is associated with a learnable referring feature, which interacts with word embeddings through cross-modal interaction~\cite{ReferSplat} to produce Gaussian--text similarity scores. 
CAM then conditions these features on camera intrinsics and extrinsics, making the similarity computation explicitly view-aware. 
During training, GCLA samples a pair of overlapping views, rasterizes the same Gaussian-level responses into two 2D masks, and jointly supervises them with paired-view losses, so that gradients from both views are coupled at shared Gaussians. 

\subsection{Camera-conditioned Alignment Modulation (CAM)}

View-consistent referring in 3D Gaussian Splatting is intrinsically challenging: the same Gaussian $G_i$ may correspond to heterogeneous visual evidence under different camera poses due to occlusion, projection distortion, and depth-dependent appearance changes. 
If camera geometry is ignored, the model is forced to explain such variations purely in semantic space, which easily entangles view-dependent evidence with view-invariant semantics and leads to semantic drift.

To explicitly model this dependency, we condition the referring field on camera geometry at the feature formation stage. 
Each calibrated camera is parameterized by intrinsics $\mathbf{K}$ and extrinsics $[\mathbf{R}|\bm{t}]$. 
We construct a pose descriptor
\begin{equation}
\bm{c} = \Gamma(\mathbf{K}, \mathbf{R}, \bm{t}) \in \mathbb{R}^{d_c},
\end{equation}
where $\Gamma(\cdot)$ concatenates the vectorized rotation $\mathrm{vec}(\mathbf{R})$, translation $\bm{t}$, and normalized intrinsic parameters (focal lengths and principal point), followed by linear normalization to ensure numerical stability. 
This descriptor is embedded via
\begin{equation}
\bm{f}_{\text{cam}} = \mathcal{E}_{\text{cam}}(\bm{c}) = \mathrm{MLP}_{\text{cam}}(\bm{c}) \in \mathbb{R}^{d},
\end{equation}
where $\mathrm{MLP}_{\text{cam}}$ is a lightweight network shared across views that maps continuous camera parameters to a coarse pose-conditioned embedding, providing view-aware cues without modeling precise camera geometry.

Given a Gaussian $G_i$ with referring feature $\bm{f}_i$ and token embeddings $\mathbf{E}=\{\bm{e}_j\}_{j=1}^{L}$, 
cross-modal interaction produces a language-aware representation
\begin{equation}
\bm{g}_i = \phi(\bm{f}_i, \mathbf{E}) \in \mathbb{R}^{d}.
\end{equation}
We then apply a camera-conditioned transform:
\begin{equation}
\tilde{\bm{g}}_i^{(v)} 
= \mathcal{T}_{\text{cam}}\!\left(\bm{g}_i, \bm{f}_{\text{cam}}^{(v)}\right)
= \bm{g}_i + \bm{f}_{\text{cam}}^{(v)},
\end{equation}
yielding a view-aware Gaussian representation for camera $v$. 
This modulation introduces a pose-dependent bias in feature space, serving as a minimal and stable way to inject geometric priors without altering the underlying semantic structure. 
It allows the same Gaussian to express view-specific evidence while preserving a shared semantic backbone across views.

The referring logit of $G_i$ under view $v$ is computed as
\begin{equation}
m_i^{(v)} 
= \psi\!\left(\tilde{\bm{g}}_i^{(v)}, \mathbf{E}\right)
= \sum_{j=1}^{L} \left(\tilde{\bm{g}}_i^{(v)}\right)^{\top} \bm{e}_j ,
\end{equation}
which explicitly makes the referring response a function of both language and camera geometry. 
Importantly, as our ablation shows, CAM alone may amplify view-dependent variations and degrade performance if not properly constrained. 
This highlights that camera-aware expressiveness must be jointly regularized by cross-view supervision, which is achieved by GCLA in the following section. 

\subsection{Gaussian-level Cross-view Logit Alignment (GCLA)}

Although CAM equips each Gaussian with a view-aware representation, view consistency cannot be guaranteed without coupling supervision across views. 
Under conventional single-view training, the learning objective provides no constraint that links the responses of the \emph{same} Gaussians observed from different cameras. 
As a result, for a Gaussian $G_i$, the predicted logits $\{ m_i^{(v)} \}$ may drift across views, making cross-view consistency an ill-posed property to emerge from optimization.

We therefore formulate cross-view consistency as a Gaussian-level alignment problem and realize it through paired-view training. 
For a given query $q$, we sample two calibrated views $(v_a, v_b)$ that share overlapping visible regions. 
Using CAM, we obtain view-conditioned logits $\{ m_i^{(v_a)} \}$ and $\{ m_i^{(v_b)} \}$ for the same underlying Gaussians. 
These logits are rasterized into predicted masks via 
\begin{equation}
\mathbf{M}_{\text{pred}}^{(v)} 
= \mathcal{R}\!\left(\{ m_i^{(v)}, \alpha_i^{(v)}, \bm{\mu}_i, \bm{\Sigma}_i \}_{i=1}^{N_v}\right),
\quad v \in \{ v_a, v_b \},
\end{equation}
where the alpha compositing naturally down-weights occluded Gaussians, making the supervision implicitly visibility-aware.

To couple the optimization of the two views, we define 
\begin{equation} \label{eq:weighted-two-view}
\footnotesize
\mathcal{L}_{\text{2view}}
=
\alpha \, \mathrm{BCE}\!\left(\mathbf{M}_{\text{pred}}^{(v_a)}, \mathbf{M}_{\text{gt}}^{(v_a)}\right)
+
(1-\alpha) \, \mathrm{BCE}\!\left(\mathbf{M}_{\text{pred}}^{(v_b)}, \mathbf{M}_{\text{gt}}^{(v_b)}\right),
\end{equation}
where $\mathbf{M}_{\text{gt}}^{(v)}$ denotes the pseudo ground-truth mask for view $v$. 
Importantly, both terms are functions of the same Gaussian parameters through $\mathcal{R}(\cdot)$. 
Thus, gradients from the two views are accumulated on shared Gaussians during back-propagation, which implicitly drives the logits to satisfy
\begin{equation}
m_i^{(v_a)} \approx m_i^{(v_b)}, 
\qquad \forall\, G_i \in \mathcal{G}_{v_a} \cap \mathcal{G}_{v_b},
\end{equation}
for Gaussians that are jointly visible in both views. 
Rather than imposing an explicit logit matching term, this formulation ties cross-view agreement to the shared optimization of Gaussian-level responses under paired-view supervision.

By turning viewpoint discrepancy into a directly optimizable objective in Gaussian space, GCLA stabilizes training under noisy pseudo supervision and complements the camera-aware expressiveness introduced by CAM. 
The overall objective is
\begin{equation}
\mathcal{L}_{\text{total}} = \lambda_1 \mathcal{L}_{\text{2view}}
+ \lambda_2 \mathcal{L}_{\text{con}},
\end{equation}
where $\lambda_1$ and $\lambda_2$ are weighting coefficients.

\begin{table}[t]
\centering
\caption{Comparison on the Ref-LERF dataset with state-of-the-art methods. 
The last row reports the relative improvement (\%) of CaRF over ReferSplat.}
\setlength{\tabcolsep}{2pt}
\label{tab:refersplat_main}
\resizebox{\linewidth}{!}{
\begin{tabular}{
l r
S[table-format=2.1, detect-weight=true, detect-inline-weight=math]
S[table-format=2.1, detect-weight=true, detect-inline-weight=math]
S[table-format=2.1, detect-weight=true, detect-inline-weight=math]
S[table-format=2.1, detect-weight=true, detect-inline-weight=math]
S[table-format=2.1, detect-weight=true, detect-inline-weight=math]
}
\toprule
\textbf{Method} & \textbf{Publisher} & \textbf{Ramen} & \textbf{Figurines} & \textbf{Teatime} & \textbf{Kitchen} & \textbf{Average} \\
\midrule
SPIn-NeRF \cite{mirzaei2023spin}      & ICCV'23 & 7.3 & 9.7 & 11.7 & 10.3 & 9.8 \\
Grounded SAM \cite{ren2024grounded}   & arXiv'24 & 14.1 & 16.0 & 16.9 & 16.2 & 15.8 \\
LangSplat \cite{Qin2024LangSplat}     & CVPR'24 & 12.0 & 17.9 & 7.6 & 17.9 & 13.9 \\
GS-Grouping \cite{Ye2024GaussianGrouping}     & ECCV'24 & 27.9 & 8.6 & 14.8 & 6.3 & 14.4 \\
GOI \cite{Qu2024GOI}                  & MM'24 & 27.1 & 16.5 & 22.9 & 15.7 & 20.6 \\
ReferSplat \cite{ReferSplat}          & ICML'25 & 28.3 & 24.3 & 27.2 & 20.1 & 25.0 \\
\rowcolor{rowblue}
CaRF (Ours)                           & -- & \textbf{33.5} & \textbf{28.7} & \textbf{29.7} & \textbf{24.7} & \textbf{29.2} \\
\rowcolor{gray!10}
\textbf{Improvement (\%)} & -- 
& +18.4 & +18.1 & +9.2 & +22.9 & +16.8 \\
\bottomrule
\end{tabular}
}
\end{table}

\section{Experiments}
\subsection{Experimental Setting}

\myPara{Datasets.}
We evaluate CaRF on three representative benchmarks: \textbf{Ref-LERF}, \textbf{LERF-OVS}, and \textbf{3D-OVS}, which collectively cover diverse challenges in 3D language grounding.  
\textbf{Ref-LERF} focuses on scene-specific referring expressions with complex spatial relations and occlusions; we adopt its official data splits and follow the confidence-weighted IoU pseudo-masking protocol introduced by ReferSplat~\cite{ReferSplat}.  
\textbf{LERF-OVS} extends 3D Gaussian Splatting to open-vocabulary segmentation across multiple scenes; we align our evaluation with LangSplat~\cite{Qin2024LangSplat} and LangSplat-V2 by querying with class-level textual phrases.  
\textbf{3D-OVS} targets large-scale, category- and room-level 3D open-vocabulary segmentation, as used in recent works such as GAGS and OpenGaussian~\cite{Wu2024OpenGaussian}.

\myPara{Evaluation Metric (mIoU).}
We adopt the mean Intersection-over-Union (\textbf{mIoU}) as the primary evaluation metric, following \cite{ReferSplat}.  
For each query–view pair, we compute the Intersection-over-Union between the predicted mask $\hat{Y}$ and the ground-truth mask $Y$ as
$
\mathrm{IoU}(Y, \hat{Y}) = \frac{|Y \cap \hat{Y}|}{|Y \cup \hat{Y}|}.
$

\begin{table}[t]
\centering
\caption{Comparison on the LERF-OVS dataset. 
The last row reports the relative improvement (\%) of CaRF over ReferSplat.}
\label{tab:refersplat_main_lerfovs}
\renewcommand{\arraystretch}{1.2}
\setlength{\tabcolsep}{2pt}
\resizebox{\linewidth}{!}{
\begin{tabular}{
l r
S[table-format=2.1, detect-weight=true, detect-inline-weight=math]
S[table-format=2.1, detect-weight=true, detect-inline-weight=math]
S[table-format=2.1, detect-weight=true, detect-inline-weight=math]
S[table-format=2.1, detect-weight=true, detect-inline-weight=math]
S[table-format=2.1, detect-weight=true, detect-inline-weight=math]
}
\toprule
\textbf{Method} & \textbf{Publisher} & \textbf{Ramen} & \textbf{Figurines} & \textbf{Teatime} & \textbf{Kitchen} & \textbf{Average} \\
\midrule
Feature-3DGS \cite{Zhou2024Feature3DGS} & CVPR'24 & 43.7 & 58.8 & 40.5 & 39.6 & 45.6 \\
LEGaussians \cite{Shi2024LEGaussians}   & CVPR'24 & 46.0 & 60.3 & 40.8 & 39.4 & 46.6 \\
LangSplat \cite{Qin2024LangSplat}       & CVPR'24 & 51.2 & 65.1 & 44.7 & 44.5 & 51.4 \\
GS-Grouping \cite{Ye2024GaussianGrouping}       & ECCV'24 & 45.5 & 60.9 & 40.0 & 38.7 & 46.3 \\
GOI \cite{Qu2024GOI}                    & MM'24 & 52.6 & 63.7 & 44.5 & 41.4 & 50.6 \\
ReferSplat \cite{ReferSplat}            & ICML'25 & 53.1 & 64.1 & 50.1 & 43.3 & 52.6 \\
\rowcolor{rowblue}
CaRF (Ours)                             & -- & \textbf{55.2} & \textbf{67.1} & \textbf{51.0} & \textbf{46.3} & \textbf{54.9} \\
\rowcolor{gray!10}
\textbf{Improvement (\%)}     & -- & +4.0 & +4.7 & +1.8 & +6.9 & +4.3 \\
\bottomrule
\end{tabular}
}
\vspace{-0.5cm}
\end{table}

\myPara{Implementation Details.}
We first pretrain an RGB-only 3DGS model and freeze its geometry parameters before learning the referring field, following standard practice for accurate visibility supervision. 
Text embeddings are extracted by BERT, and Gaussian--text interaction follows ReferSplat~\cite{ReferSplat}. 
During training, we sample paired views $(v_a, v_b)$ with at least $30\%$ overlap and use $\alpha=0.5$ in \cref{eq:weighted-two-view}. 
Pseudo masks are generated from $K$ SAM candidates via confidence-weighted IoU aggregation. 
We optimize the model with Adam for 30k iterations, using learning rates of $2.5\times10^{-3}$ for the referring field and contrastive head, and $1\times10^{-4}$ for the camera MLP. 
The feature dimension is $d=128$, with mixed-precision training and gradient clipping of $1.0$. 
All experiments are conducted on RTX A6000 GPUs. We reproduce ReferSplat under the same settings, and results are averaged over five runs with $\lambda_1=\lambda_2=1$ for balanced weighting.

\subsection{Results on the Ref-LERF Dataset}
As shown in \cref{tab:refersplat_main}, CaRF achieves new state-of-the-art performance on Ref-LERF, improving the average mIoU from 25.0 (ReferSplat) to 29.2 (+16.8\%). 
Consistent gains are observed across all scenes, including Ramen (+18.4\%), Figurines (+18.1\%), Teatime (+9.2\%), and Kitchen (+22.9\%). 
The largest improvements appear in \textit{Kitchen} and \textit{Ramen}, where heavy occlusion, clutter, and fine structures make single-view pseudo supervision particularly brittle. 
Overall, these results demonstrate that camera-aware cross-view alignment substantially stabilizes multi-view reasoning for referring 3DGS.

\begin{table}[t]
\centering
\caption{Comparison on the 3D-OVS dataset. The last row reports the relative improvement (\%) of CaRF over ReferSplat.}
\label{tab:refersplat_main_3dovs}
\setlength{\tabcolsep}{2pt}
\resizebox{\linewidth}{!}{
\begin{tabular}{
l r
S[table-format=2.1, detect-weight=true, detect-inline-weight=math]
S[table-format=2.1, detect-weight=true, detect-inline-weight=math]
S[table-format=2.1, detect-weight=true, detect-inline-weight=math]
S[table-format=2.1, detect-weight=true, detect-inline-weight=math]
S[table-format=2.1, detect-weight=true, detect-inline-weight=math]
S[table-format=2.1, detect-weight=true, detect-inline-weight=math]
}
\toprule
\textbf{Method} & \textbf{Publisher} & \textbf{Bed} & \textbf{Bench} & \textbf{Room} & \textbf{Sofa} & \textbf{Lawn} & \textbf{Average} \\
\midrule
Feature-3DGS \cite{Zhou2024Feature3DGS}  & CVPR 2024  & 83.5 & 90.7 & 84.7 & 86.9 & 93.4 & 87.8 \\
LEGaussians \cite{Shi2024LEGaussians}    & CVPR 2024  & 84.9 & 91.1 & 86.0 & 87.8 & 92.5 & 88.5 \\
LangSplat \cite{Qin2024LangSplat}         & CVPR 2024  & 92.5 & 94.2 & 94.1 & 90.0 & 96.1 & 93.4 \\
GS-Grouping \cite{Ye2024GaussianGrouping}         & ECCV'24  & 83.0 & 91.5 & 85.9 & 87.3 & 90.6 & 87.7 \\
GOI \cite{Qu2024GOI}                      & MM'24 & 89.4 & 92.8 & 91.3 & 85.6 & 94.1 & 90.6 \\
ReferSplat \cite{ReferSplat}              & ICML'25   & 90.2 & 93.8 & 94.1 & 90.8 & 95.5 & 92.9 \\
\rowcolor{rowblue}
CaRF (Ours)                                & --         & \textbf{92.1} & \textbf{94.2} & \textbf{96.8} & \textbf{93.2} & \textbf{97.3} & \textbf{94.7} \\
\rowcolor{gray!10}
\textbf{Improvement (\%)}        & --         & +2.1 & +0.4 & +2.9 & +2.6 & +1.9 & +2.0 \\
\bottomrule
\end{tabular}
}
\end{table}

\subsection{Results on 3D Open-Vocabulary Segmentation Datasets}

As shown in \cref{tab:refersplat_main_lerfovs,tab:refersplat_main_3dovs}, CaRF consistently outperforms prior methods on both \textbf{LERF-OVS} and \textbf{3D-OVS}. 
On \textbf{LERF-OVS}, CaRF achieves 54.9 mIoU, surpassing ReferSplat by 4.3\%, with larger gains in cluttered scenes such as \textit{Kitchen} (+6.9\%) and \textit{Figurines} (+4.7\%), where viewpoint drift is more severe. 
On \textbf{3D-OVS}, CaRF reaches 94.7 mIoU, improving over ReferSplat by 2.0\%; the smaller margin is expected since the baseline is already near-saturated, leaving limited room for improvement. 
Overall, these results demonstrate that CaRF enables robust open-vocabulary segmentation across scenes.

\begin{table}[t]
\centering
\caption{Unified ablation study of CaRF on Ref-LERF.}
\label{tab:ablation_all}
\setlength{\tabcolsep}{2pt}
\resizebox{\linewidth}{!}{
\begin{tabular}{lccccccc}
\toprule
\textbf{Setting} & \textbf{GCLA} & \textbf{CAM} & \textbf{Selection} & \textbf{Cam Fusion} & \textbf{\#Views} & \textbf{Ramen} & \textbf{Kitchen} \\
\midrule
Baseline (ReferSplat) & \xmark & \xmark & LERF & -- & 1 & 28.3 & 20.1 \\
+ GCLA              & \cmark & \xmark & LERF & -- & 2 & 31.6 & 22.4 \\
+ CAM               & \xmark & \cmark & LERF & MLP & 1 & 24.3 & 13.5 \\
\midrule
Ours (Cosine)        & \cmark & \cmark & Cosine & MLP & 2 & 33.5 & 24.7 \\
Ours (LERF)          & \cmark & \cmark & LERF   & MLP & 2 & 31.2 & 23.2 \\
\midrule
Ours (Post-fusion)   & \cmark & \cmark & Cosine & Post & 2 & 25.6 & 18.3 \\
Ours (Lang-enc)      & \cmark & \cmark & Cosine & Lang & 2 & 28.3 & 22.4 \\
\midrule
Ours (3-view)        & \cmark & \cmark & Cosine & MLP & 3 & 33.7 & 23.1 \\
Ours (4-view)        & \cmark & \cmark & Cosine & MLP & 4 & 32.4 & 24.1 \\
\rowcolor{rowblue}
Ours (2-view)        & \cmark & \cmark & Cosine & MLP & 2 & 33.5 & 24.7 \\
\bottomrule
\end{tabular}
}
\vspace{-0.5cm}
\end{table}

\subsection{Ablation Study}

We conduct a unified ablation study on Ref-LERF to analyze key design choices of CaRF, including paired-view supervision (GCLA), camera-aware encoding (CAM), Gaussian selection, camera fusion design, and the number of training views. 
As summarized in \cref{tab:ablation_all}, introducing \textbf{GCLA} brings consistent gains over the baseline, while using \textbf{CAM} alone degrades performance, indicating that camera-conditioned features require cross-view constraints to form stable correspondences. 
Combining both yields the best results, confirming their complementarity.  
For Gaussian selection, cosine similarity outperforms LERF-style relevancy scoring, providing more stable and accurate responses. 
Among camera fusion strategies, the MLP-based design achieves the highest performance, while post-fusion and language-level fusion lead to clear drops, showing that directly conditioning Gaussian features is most effective. 
Finally, using two views offers the best accuracy–efficiency trade-off, as more views bring marginal gains but higher cost.  
Overall, these results validate that CaRF’s improvements stem from jointly modeling camera geometry and Gaussian-level cross-view alignment.

\subsection{Qualitative and Quantitative Analysis}

\begin{figure}
    \centering
    \includegraphics[width=\linewidth]{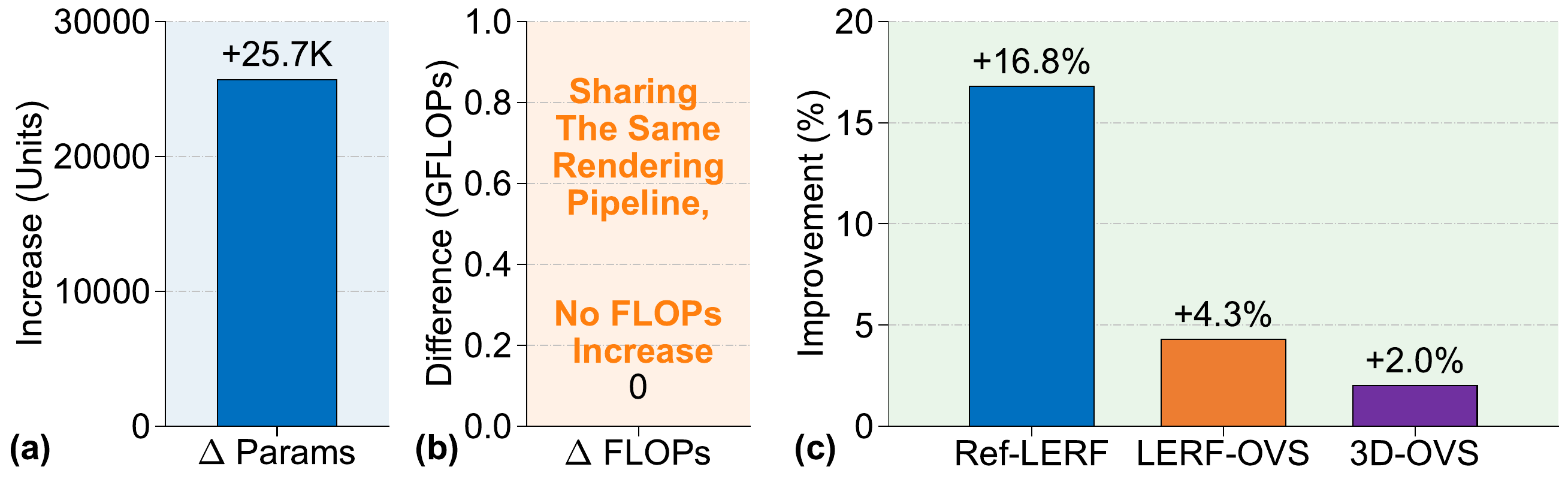}
    \caption{
Performance--computation comparison between CaRF and ReferSplat \cite{ReferSplat}. 
(a) Parameter increase ($\Delta$Params). 
(b) FLOPs difference ($\Delta$FLOPs). 
(c) Performance improvement ($\Delta$mIoU) across three benchmarks.
}

    \label{fig:delta_summary}
    \vspace{-0.35cm}
\end{figure}

\textbf{Balance Between Computation and Accuracy}
As shown in \cref{fig:delta_summary}, CaRF achieves a favorable accuracy--efficiency trade-off. 
The only extra parameters come from the lightweight CAM MLP, adding merely 25.7K parameters to map camera poses to a 128-d embedding. 
The cross-view alignment loss (GCLA) is used only during training and introduces moderate overhead (approximately $2\times$ training time with negligible VRAM increase due to dual-view rasterization), while inference follows the rendering pipeline of ReferSplat with almost no additional FLOPs.
Despite this small training overhead, CaRF improves mIoU by 16.8\%, 4.3\%, and 2.0\% on Ref-LERF, LERF-OVS, and 3D-OVS, respectively, demonstrating that camera-aware alignment significantly enhances 3D language grounding while keeping inference lightweight.

\begin{figure}
    \centering
    \includegraphics[width=\linewidth]{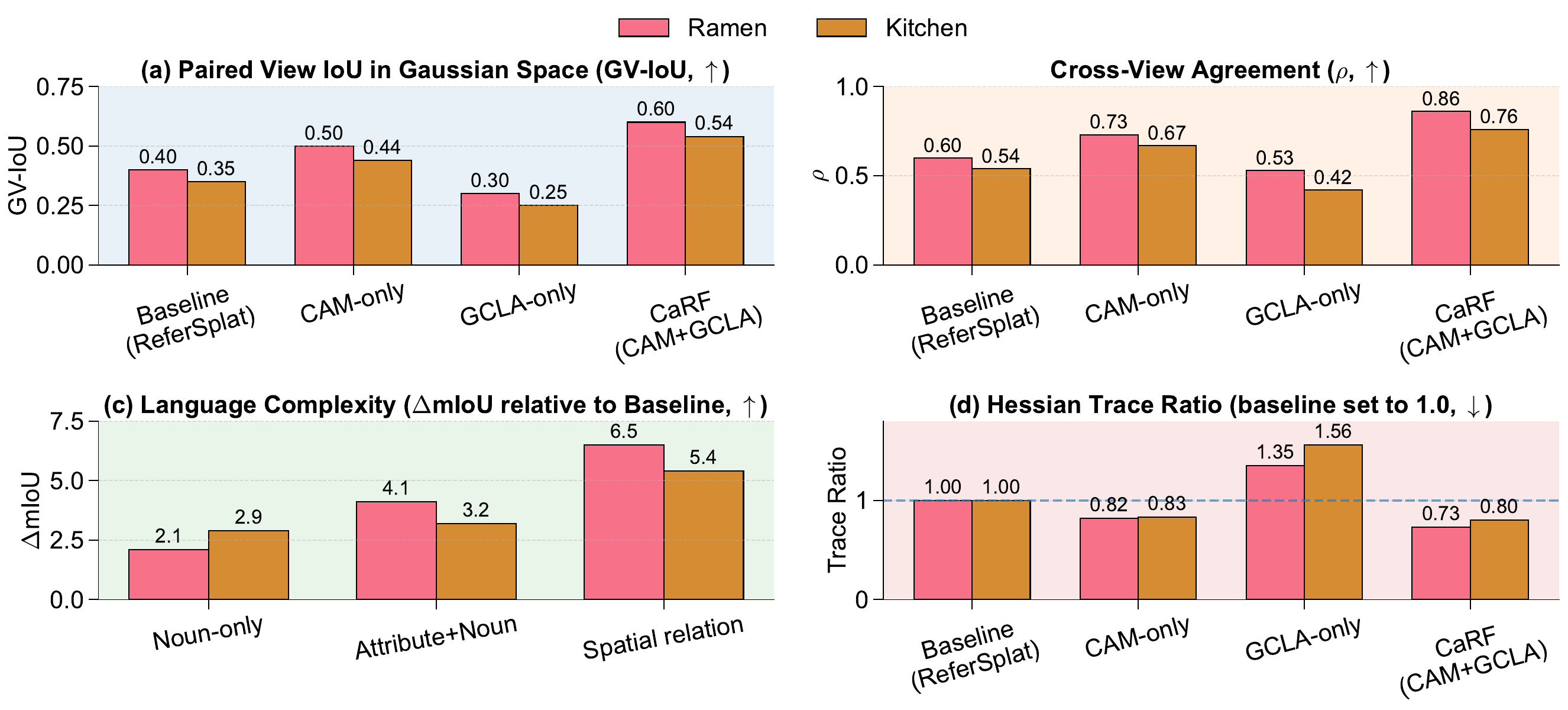}
    \caption{
Analysis of CaRF on Ref-LERF. We report (a) GV-IoU, (b) cross-view agreement, (c) gains under increasing language complexity, and (d) Hessian trace ratio (lower is better) on \textit{Ramen} and \textit{Kitchen}. 
}
    \label{fig:multi}
    \phantomsubcaption\label{fig:multi-a}
\phantomsubcaption\label{fig:multi-b}
\phantomsubcaption\label{fig:multi-c}
\phantomsubcaption\label{fig:multi-d}
\vspace{-0.5cm}
\end{figure}

\begin{figure}[t]
\centering
\includegraphics[width=1\linewidth,clip,trim=85 10 150 10]{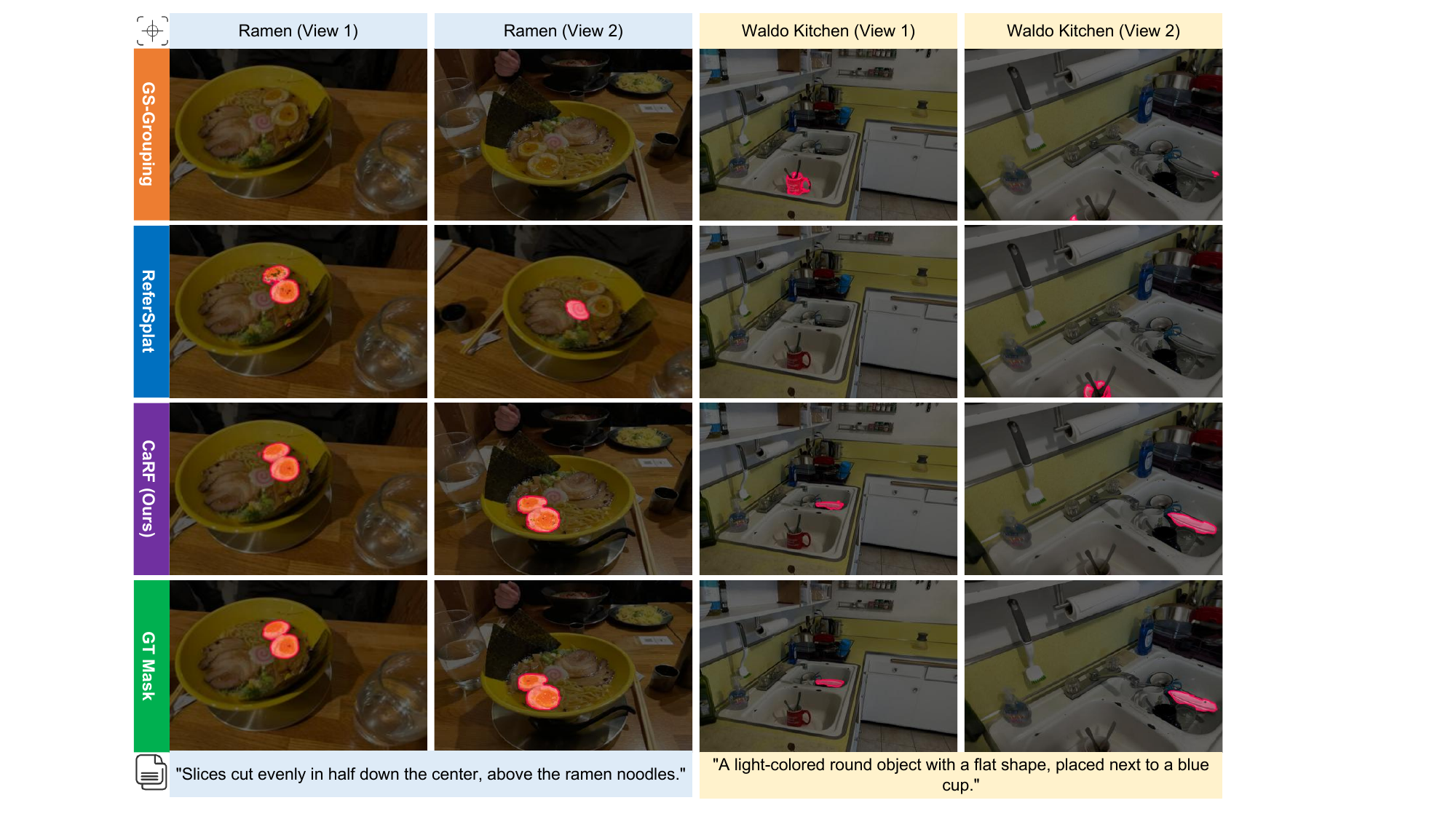}
\caption{
Qualitative comparisons on Ref-LERF across two scenes and two views: compared with \textbf{\textcolor[HTML]{ED7D31}{GS-Grouping}}~\cite{ye2024gaussian} and \textbf{\textcolor[HTML]{0070C0}{ReferSplat}}~\cite{ReferSplat}, our \textbf{\textcolor[HTML]{7030A0}{CaRF}} produces view-consistent masks that closely match the \textbf{\textcolor[HTML]{00B050}{GT}}.
}

\label{fig:vis}
\vspace{-0.35cm}
\end{figure}

\begin{figure}[t]
\centering
\includegraphics[width=1\linewidth,clip,trim=85 130 150 10]{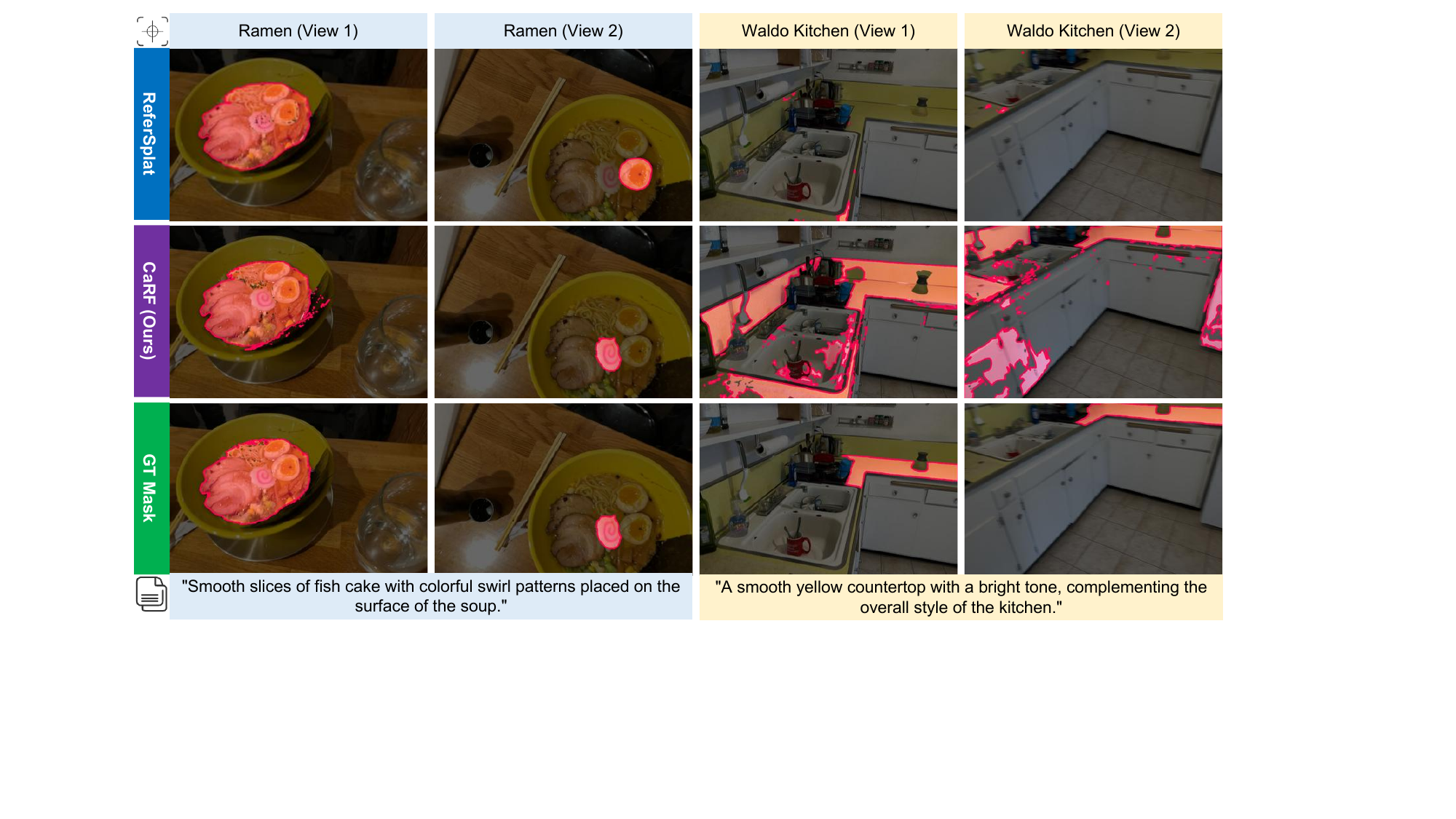}
\caption{
Failure cases on the Ref-LERF dataset. 
}
\label{fig:vis2}
\vspace{-0.5cm}
\end{figure}

\textbf{Cross-View Consistency.} 
In \cref{fig:multi-a}, we report the Paired-View IoU in Gaussian space, $\mathrm{GV\text{-}IoU}=\frac{|S^{(v_a)}\cap S^{(v_b)}|}{|S^{(v_a)}\cup S^{(v_b)}|}$, where  
$S^{(v)}=\{i \mid m_i^{(v)} \text{ in top-}k\}$,
which measures the overlap of foreground Gaussians predicted from two calibrated views. 
CaRF achieves the highest GV-IoU, reaching $0.60$ on \textit{Ramen} and $0.54$ on \textit{Kitchen}, surpassing ReferSplat ($0.40/0.35$).

\textbf{Cross-View Agreement.} 
In \cref{fig:multi-b}, we compute the Pearson correlation 
$\rho=\mathrm{Corr}(\bm{m}^{(v_a)},\bm{m}^{(v_b)})$ between per-Gaussian logits rendered from paired views to assess continuous response consistency. 
CaRF again performs best with $\rho=0.86$ on \textit{Ramen} and $0.76$ on \textit{Kitchen}.

\textbf{Language Complexity.} 
In \cref{fig:multi-c}, we group queries into noun-only, attribute plus noun, and spatial relation, and report the mIoU gains over ReferSplat. 
CaRF shows increasingly larger improvements as language becomes more complex, achieving up to $+6.5$ and $+5.4$ mIoU for spatial queries.

\textbf{Optimization Stability.} 
In \cref{fig:multi-d}, we estimate the Hessian trace ratio of the training loss using Hutchinson approximation, normalized by ReferSplat. 
CaRF yields the lowest ratios, 0.73 on \textit{Ramen} and 0.80 on \textit{Kitchen}, indicating the most stable optimization among all variants.

\textbf{Case Study.} 
In \cref{fig:vis}, CaRF produces more accurate and complete masks that better match the referring expressions, while Gaussian Grouping often confuses nearby regions due to category-level clustering, and ReferSplat exhibits view-inconsistent or fragmented predictions. 
Failure cases in \cref{fig:vis2} mainly stem from noisy or ambiguous pseudo masks rather than model errors. 
Even under such imperfect supervision, CaRF still yields smoother and more consistent segmentations than ReferSplat, demonstrating robustness to label noise.

\section{Conclusion and Discussion}

We presented CaRF, a camera-aware R3DGS method that tackles multi-view inconsistency. 
By combining camera-conditioned alignment modulation with Gaussian-level cross-view logit alignment, CaRF enables geometry-aware and view-consistent language grounding directly in Gaussian space. 
Experiments demonstrate consistent mIoU gains over prior methods, establishing CaRF as a strong baseline for R3DGS and highlighting the importance of coupling language grounding with camera geometry for robust 3D understanding.

CaRF still has limitations. It relies on pseudo masks for supervision and requires per-scene optimization, which may limit scalability to dynamic scenes. 
Future work will explore stronger 3D priors and more lightweight designs to extend CaRF toward dynamic 3D perception.


{\small
\bibliographystyle{ieeetr}
\bibliography{cvmbib}
}

\end{document}